\documentclass[english]{llncs}
\usepackage[T1]{fontenc}
\usepackage[latin9]{inputenc}
\usepackage{xcolor}
\usepackage{booktabs}
\usepackage{graphicx}
\usepackage[numbers]{natbib}
\PassOptionsToPackage{normalem}{ulem}
\usepackage{ulem}

\makeatletter

\providecommand{\tabularnewline}{\\}
\providecolor{lyxadded}{rgb}{0,0,1}
\providecolor{lyxdeleted}{rgb}{1,0,0}
\DeclareRobustCommand{\mklyxadded}[1]{\textcolor{lyxadded}\bgroup#1\egroup}
\DeclareRobustCommand{\mklyxdeleted}[1]{\textcolor{lyxdeleted}\bgroup\mklyxsout{#1}\egroup}
\DeclareRobustCommand{\mklyxsout}[1]{\ifx\\#1\else\sout{#1}\fi}

\@ifundefined{date}{}{\date{}}
\makeatother

\usepackage{babel}
\begin{document}
\title{Comparing GenAI to Machine Learning and Manual Review for eDiscovery}
\author{Gordon V. Cormack \and Maura R. Grossman}
\institute{University of Waterloo, ON, Canada}
\authorrunning{Cormack, Grossman, Pickens}
\titlerunning{Comparing GenAI to ML and Human Review}
\maketitle
\begin{abstract}
Keeling et al. \citep{Keeling2026Generative} report on a case study
recently conducted by Redgrave LLP in which they assert that Relativity's
Generative AI Review tool (aiR) achieved 88\% recall while Relativity
Assisted Learning (RAL), its machine-learning tool, achieved 64\%
recall. These results are not comparable because the methods employed
have different objectives, because of inter-assessor disagreement,
and because they represent different recall-precision and cost-benefit
trade-offs. Our analysis of the Redgrave data indicates that, when
these differences are taken into account, aiR and manual review are
comparable. Furthermore, established machine-learning models employing
Continuous Active Learning (CAL) and Scalable Continuous Active Learning
(S-CAL) show state-of-the-art effectiveness, and superior efficiency,
when trained only on a sample of 2,000 first-pass-reviewer responsiveness
determinations. 
\end{abstract}

\section{Introduction}

Keeling et al. \citep{Keeling2026Generative} conducted a head-to-head
case study for Redgrave LLP (the Redgrave Case Study) comparing Relativity
aiR Review (aiR), a generative AI tool, with an established managed
review process employing Relativity Assisted Learning (RAL) \citep{Keeling2026Generative}.
Keeling et al. reported that Relativity's aiR achieved 87.7\% recall
and 28.6\% precision, compared to RAL's 64.4\% recall and 39.2\% precision.

These results -- reported in multiple headlines -- are incomparable
for three principal reasons:
\begin{enumerate}
\item The RAL process identified for production those documents that were
identified by RAL \emph{and also coded responsive} by a team of junior
``contract lawyers'' (the L1 team). In contrast, the aiR process
identified for production only those documents that were identified
by aiR, \emph{without such human review}. We call these methods (L1-)
``filtered'' and ``unfiltered'' TAR methods, respectively.
\item The L1 review team, the aiR prompt engineer, and the subject matter
expert (SME), whose assessments were used for evaluation, have different
conceptions of responsiveness, which will naturally influence recall
and precision estimates; results according to the SME may simply reflect
how similar the L1 and prompt engineer teams' conceptions of responsiveness
were to the SME's, rather than a technological difference.
\item The aiR and RAL results represent different recall-precision trade-offs.
Choosing the best trade-off involves balancing proportionality factors
such as completeness, cost, burden, risk, and timeliness. We cannot
automatically say that one production is ``better'' than another
without considering the risk and burden to both the producing and
requesting parties, respectively.
\end{enumerate}
This work shows that when the L1 review team has the final say on
which documents are responsive (filtered TAR), or when all documents
identified by the system are deemed responsive regardless of the L1
coding (unfiltered TAR), the differences in review effectiveness between
aiR and managed review using RAL essentially vanish. Furthermore,
decade-old supervised machine-learning protocols -- e.g., Continuous
Active Learning (CAL) \citep{cormackEvaluationMachinelearningProtocols2014}
for filtered TAR, and Scalable Continuous Active Learning (S-CAL)
\citep{cormackScalabilityContinuousActive2016} for unfiltered TAR
-- yield equal or better efficiency and effectiveness.

\section{Dataset}

The Redgrave Case Study used 45,004 documents from the UCSF Opioid
Industry Documents Archive \citep{IndustryDocumentsLibrary} (the
Redgrave Dataset). RAL managed review involved manual coding for responsiveness
of 34,409 documents in the order identified by RAL, followed later
by manual coding of an additional 10,595 documents not identified
by RAL, resulting in L1 coding for the entire Redgrave Dataset. The
aiR review involved composing GenAI prompts, and having the GenAI
label each document. Independent review by an SME of a simple random
sample (n=1,000) identified 73 documents as responsive. Treating the
SME coding as ground truth, the RAL review achieved 64.4\% recall
(95\% CI: 52.3\%--75.3\%) and 39.2\% precision (30.4\%--48.4\%),
while aiR achieved 87.7\% recall (95\% CI: 77.9\%--94.2\%) and 28.6\%
precision (22.8\%--35.0\%).

\section{Filtered and Unfiltered TAR Comparisons}

\begin{figure}
\begin{centering}
\includegraphics[width=0.9\columnwidth]{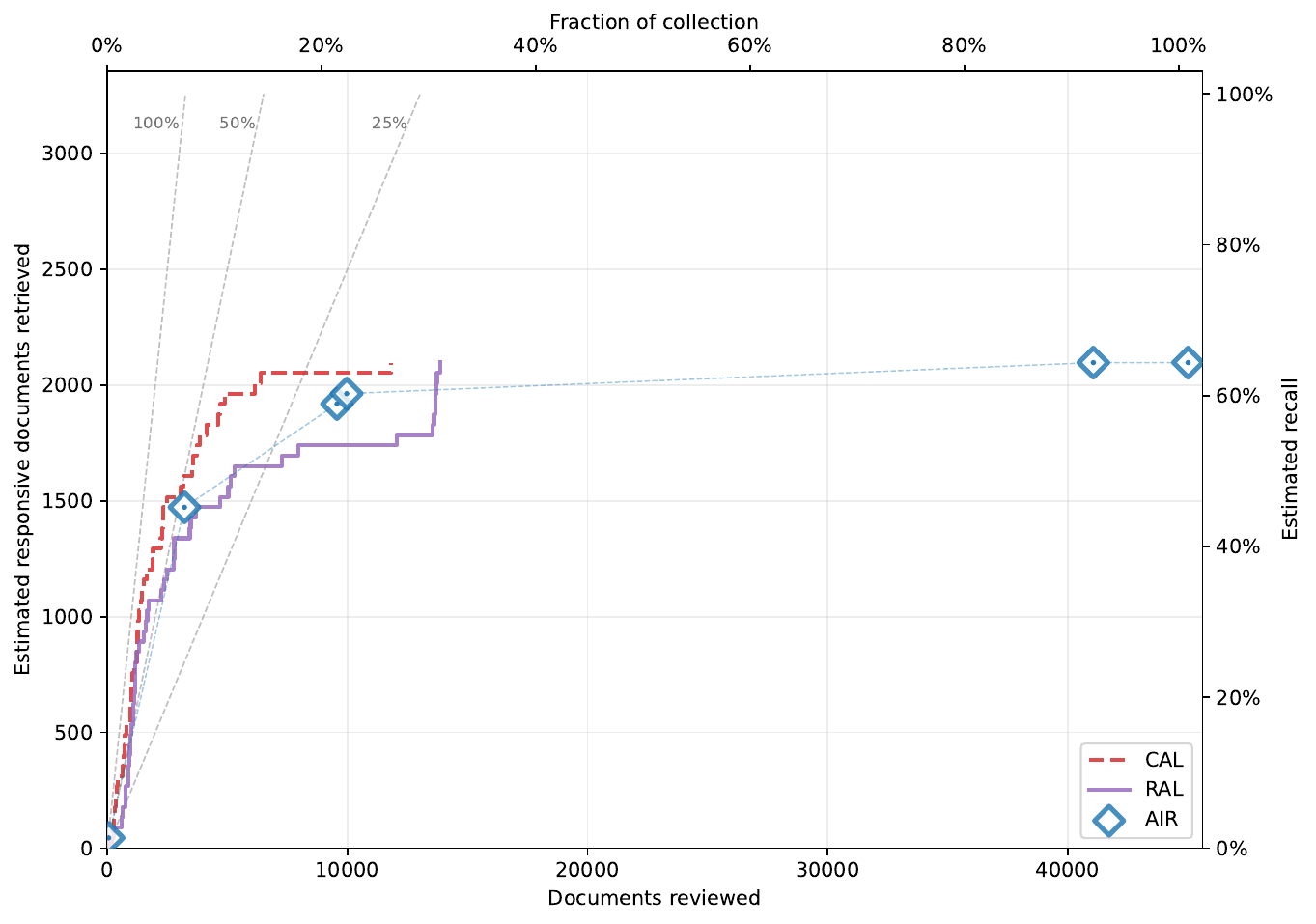}
\par\end{centering}
\begin{centering}
\caption{\label{fig:Responsive-filtered}Responsive documents identified by
aiR, RAL, and CAL, filtered by L1 responsiveness coding.}
\par\end{centering}
\begin{centering}
\includegraphics[width=0.9\columnwidth]{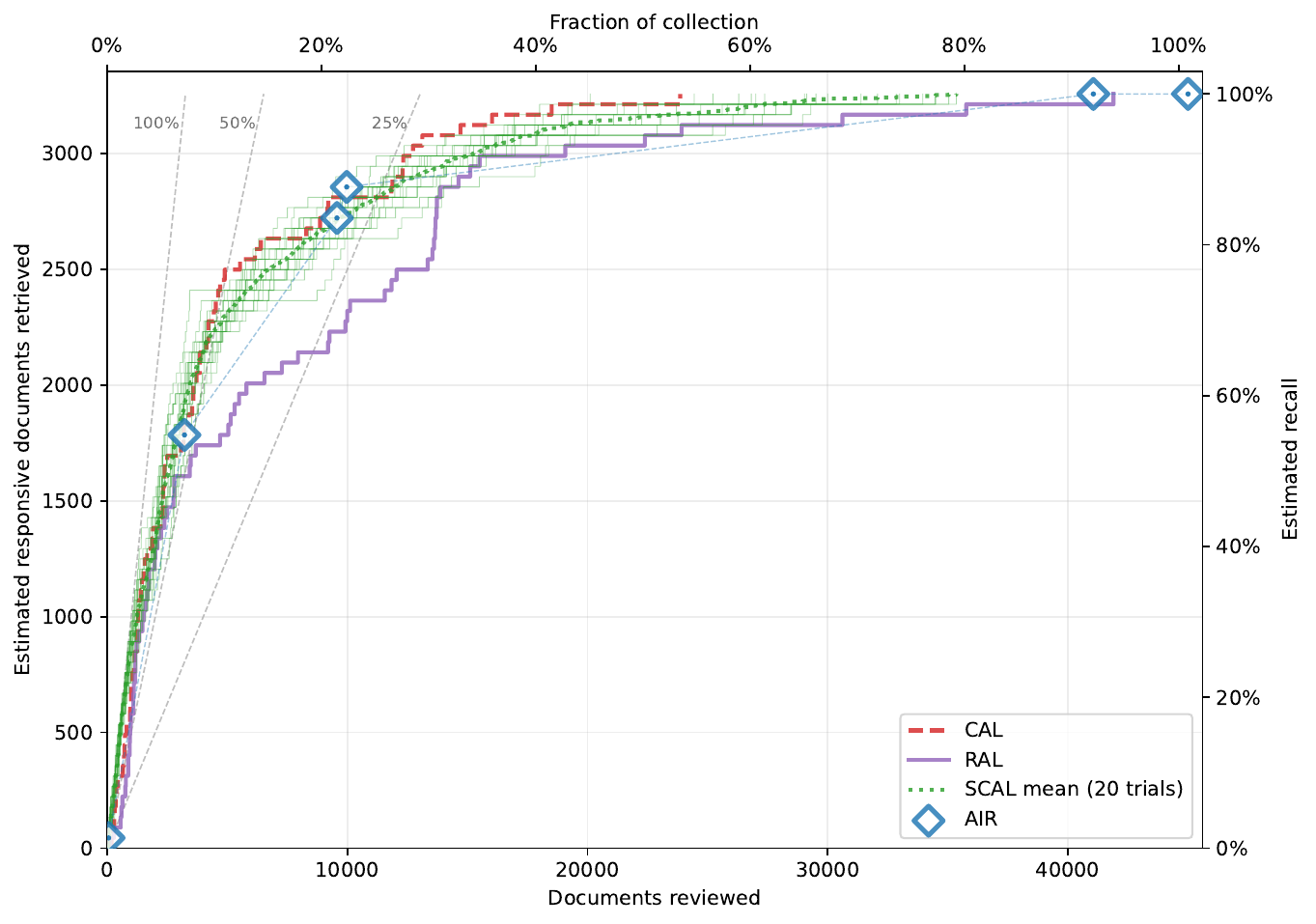}
\par\end{centering}
\centering{}\caption{\label{fig:Responsive-unfiltered}Responsive documents identified
by aiR, RAL, CAL, and S-CAL, unfiltered by L1 responsiveness coding.
Thin S-CAL curves show 20 individual trials; the thick curve shows
their mean.}
\end{figure}
Also treating the SME coding as ground truth, we plotted gain curves
for the filtered (Figure \ref{fig:Responsive-filtered}) aiR, RAL,
and CAL methods, and for the unfiltered (Figure \ref{fig:Responsive-unfiltered})
aiR, RAL, CAL, and S-CAL methods. The CAL and S-CAL implementations
were derived from the open-source Baseline Model Implementation (BMI)\footnote{plg.uwaterloo.ca/\textasciitilde gvcormac/trecvm/; implementation
extracted from plg.uwaterloo.ca/\textasciitilde gvcormac/trecvm/vmscripts.tar.} used for the TREC 2015 and 2016 Total Recall Tracks \citep{roegiest2015total,grossman2016total}.
The CAL version was modified to retrain after each L1 coding decision,
rather than in progressively larger batches (per Ghelani et al. \citep{ghelani2018refresh}),
and the S-CAL version was modified to require L1 coding for only a
weighted statistical sample of 2,000 documents from the Redgrave Dataset,
excluding the SME-coded sample of 1,000 documents.

\section{Assessor Agreement and Disagreement}

Recall and precision are measured against a particular assessor's
judgments, not an infallible ground truth. In this case, the L1 reviewers
coded as non-responsive 34\% of the documents the SME deemed responsive,
while the SME coded as non-responsive 61\% of those that the L1 reviewers
deemed responsive.

The agreement and disagreement between assessors is typically quantified
using ``overlap,'' also known as Jaccard index: Overlap is the number
of documents that a pair of assessors both label responsive, divided
by the number that one or both label responsive.
\begin{table}
\centering
\caption{\label{tab:overlapredgrave}Inter-assessor overlap (Jaccard index)
in the Redgrave Case Study.}

\begin{tabular}{lc}
SME \& L1 & 0.322\tabularnewline
SME \& S-CAL & 0.361\tabularnewline
SME \& aiR (liberal) & 0.275\tabularnewline
SME \& aiR (strict) & 0.421\tabularnewline
\end{tabular}
\end{table}
\begin{table}
\centering
\caption{\label{tab:overlaplit}Intra- and Inter-assessor overlap (Jaccard
index) reported in the literature.}

\begin{tabular}[t]{lc}
\multicolumn{2}{c}{Retired Analysts \citep{webberAlternativeAssessorDisagreement2012}}\tabularnewline
\midrule
Primary \& A & 0.295\tabularnewline
Primary \& B & 0.350\tabularnewline
 & \ \tabularnewline
\end{tabular}\thickspace{}%
\begin{tabular}[t]{lc}
\multicolumn{2}{c}{DOJ Second Request \citep{roitblatDocumentCategorizationLegal2010}}\tabularnewline
\midrule
Original \& Team A & 0.163\tabularnewline
Original \& Team B & 0.158\tabularnewline
 & \ \tabularnewline
\end{tabular}\thickspace{}%
\begin{tabular}[t]{lc}
\multicolumn{2}{c}{State Archivist \citep{cormackNavigatingImprecisionRelevance2017a}}\tabularnewline
\midrule
Roger I \& II (legal hold) & 0.806\tabularnewline
Roger I \& II (public record) & 0.602\tabularnewline
Roger I \& II (restricted record) & 0.642\tabularnewline
\end{tabular}
\end{table}

Table \ref{tab:overlapredgrave} shows the overlap between SME coding
and L1, S-CAL, and two aiR interpretations: \emph{liberal} (including
``marginal'' and ``error'' categories, as necessary to achieve
nearly 88\% recall) and \emph{strict }(including only ``responsive''
and ``highly responsive'' categories). The liberal and strict interpretations
correspond to the second and fourth diamonds, respectively, in Figures
\ref{fig:Responsive-filtered} and \ref{fig:Responsive-unfiltered}.
S-CAL overlap is calculated at its self-estimated recall-precision
crossover.

Table \ref{tab:overlaplit} shows that reviewer disagreement is not
unusual, ranging from 0.158 to 0.350 between professional reviewers,
and from 0.602 to 0.806 between judgments made by the same SME at
different times. Disagreement does not imply that either review is
unreasonable. Managed L1 review is widely accepted as a reasonable
basis for production decisions in the legal industry. Independent
assessments, whether by L1 reviewers or SMEs, can support comparisons
of retrieval methods performing the same task \citep{voorheesVariationsRelevanceJudgments2000}.
They do not establish an infallible ground truth.

The differences between results for the filtered and unfiltered tasks
-- both gain curves and recall-precision estimates -- are largely
explainable by inter-assessor disagreement. Within each task, the
continuous gain curves come within the sampling margin of error of
the discrete aiR results, and the remaining differences -- whether
due to chance, training, or method -- have little practical significance.

\section{\label{sec:Proportionality-and-the}Proportionality and the Recall-Precision
Trade-off}

The doctrine of proportionality occasions us to balance the burden
of reviewing or producing more documents, with their value. We can
inform this analysis by examining the slope of the gain curve, if
it is known. For RAL and CAL, the gain curve is constructed incrementally,
and the utility of proceeding can be assessed by the frequency with
which responsive documents are identified (i.e., the estimated slope
of the gain curve). From S-CAL's self-estimate of the entire gain
curve, we may identify the point of diminishing returns. For aiR,
we have only two viable cut points -- liberal or strict -- and no
principled way to choose between them.

\section{Limitations}

The Redgrave Case Study considers a single matter, a single L1 review,
and a single prompt engineering effort.

The Redgrave Dataset is a small, non-random subset of the 1.4M document
Mallinckrodt collection, with an estimated richness of 7.3\%, which
may not be representative of the richness or nature of the documents
in the larger collection (or other collections).

The opioids litigation has garnered a large amount of press and commentary,
which may limit the applicability of results derived from it -- or
any similarly notorious dataset -- to novel matters that are not
equally publicized.

Redgrave's aiR prompts and the methodology to develop them remain
confidential.

Opaque aiR pricing precludes evaluation of cost in the proportionality
analysis. Relativity documentation states that a subscription includes
an allowance of AI \textquotedblleft units\textquotedblright{} and
that overage charges apply. When asked, Relativity declined to quantify
the allowance or the overage charges.

\section{Discussion and Conclusions}

aiR can outperform RAL at specific points on the gain curve. But at
the actual cutoff values -- when comparing the methods using a common
task -- aiR does \emph{not} outperform RAL. In the filtered TAR task,
RAL achieves 64\% recall (the highest possible given SME-L1 disagreement),
while aiR achieves 59.5\%. \emph{CAL is superior to both at all points
in the gain curve.} In the unfiltered TAR task, aiR achieves 88\%
recall, while RAL achieves 97.2\%.

Using RAL (or CAL) for unfiltered TAR review would most likely be
considered disproportionate because of review cost. S-CAL achieves
a similar result with a fixed review budget -- in this case, 2,000
documents. S-CAL dominates RAL, and substantially matches both aiR-liberal
and aiR-strict points. S-CAL, like RAL, exceeds 90\% recall and eventually
achieves 100\% recall, which aiR never reaches.

At its self-calibrated recall-precision crossover, S-CAL agrees more
closely with the SME than its L1 teacher does, consistent with prior
results \citep{cormackUnbiasedValidationTechnologyAssisted2024}.
Self-calibration may be used to select other points on the gain curve,
as occasioned by proportionality factors \citep{cormackScalabilityContinuousActive2016}.

To assess scalability from the Redgrave Dataset to the full 1.4M-document
Mallinckrodt collection, we consider cost and effectiveness. CAL and
RAL review costs depend primarily on the number of responsive documents,
not the collection size. Adding mostly non-responsive documents would
have little effect, but at a similar richness, review effort could
increase by an order of magnitude.

On the other hand, aiR prompt engineering and S-CAL training are sunk
costs, independent of collection size or the number of responsive
documents. We have no idea of the cost to run aiR over 1.4M documents;
S-CAL can be trained and run on commodity hardware in 8 seconds elapsed
time and 32 seconds CPU time, once the documents have been vectorized,
which takes less than one hour.

Keeling et al. characterize 7.3\% responsive as ``low richness (7\%,
well below the 20--25\% design target)'' and suggest that with higher
richness the methods tested would have achieved superior precision.
While we find no guidance or empirical result supporting these assertions,
we observe that, if they were true, the converse would hold: The methods
tested would achieve an inferior result on the full corpus because
of its lower richness. We note that the richness of the TREC 2009
Legal track topics, which were used to establish the effectiveness
and efficiency of TAR \citep{grossmanTechnologyAssistedReviewEDiscovery2011},
ranged in richness from 0.3\% to 1.5\% (median 0.6\%) and achieved
precision ranging from 67.3\% (with 86.5\% recall) to 88.4\% (with
69.2\% recall), in a collection of 569,034 emails.

For this particular task, the Redgrave Case Study shows aiR to be
comparable to, but not superior to RAL, whether the tool or the L1
reviewer makes the final determination of responsiveness. When the
L1 reviewer makes the final determination, CAL achieves the same recall
as RAL, with less review effort. When the tool makes the final determination,
S-CAL, trained on 2,000 L1 reviewer responsiveness determinations,
is as effective as aiR, offers a calibrated adjustment of the recall-precision
trade-off, and runs quickly and cheaply on commodity hardware. aiR
offers only two viable recall-precision choices, offers no calibrated
means to choose, and runs slowly on specialized hardware.

\section{Acknowledgement}

We thank Redgrave LLP for making the case-study data available for
independent research and for their assistance in understanding the
data and study methodology.

\bibliographystyle{splncs04}
\bibliography{references}

\end{document}